\documentclass[runningheads]{llncs}
\usepackage{graphicx}
\usepackage{amsmath,amssymb} 
\usepackage{color}
\usepackage{subfigure}
\usepackage{booktabs} 
\usepackage{amsmath}
\usepackage{amssymb}
\usepackage{multirow}
\usepackage{algorithm}
\usepackage[noend]{algpseudocode}
\usepackage{algorithmicx}

\usepackage[pagebackref=true,breaklinks=true,letterpaper=true,colorlinks,bookmarks=false]{hyperref}
\usepackage[usenames,dvipsnames,svgnames,table]{xcolor}

\DeclareMathOperator*{\argmin}{arg\,min}

\newcommand{\etal}{\textit{et al}.}
\newcommand{\ie}{\textit{i}.\textit{e}.}
\newcommand{\eg}{\textit{e}.\textit{g}.}
\def\Vec#1{{\boldsymbol{#1}}}
\def\Mat#1{{\mathbf{#1}}}

\begin{document}

\title{Scalable Deep $k$-Subspace Clustering} 
\titlerunning{Scalable Deep $k$-Subspace Clustering} 


\author{Tong Zhang\inst{1,5} \and
Pan Ji\inst{2,4} \and
Mehrtash  Harandi\inst{3,5} \and \\  
Richard Hartley\inst{1,5} \and  
Ian Reid\inst{2,5} 
}
%

\authorrunning{T. Zhang et al.} 


\institute{$^1$ Australian National University, $^2$ University of Adelaide, $^3$ Monash University, \\$^4$ NEC Labs America, $^5$ Australian Centre for Robotic Vision
}

\maketitle

\begin{abstract}
Subspace clustering algorithms are notorious for their scalability issues
because building and processing large affinity matrices are demanding.
In this paper, we introduce a method that simultaneously learns an embedding space 
along subspaces within it to minimize a notion of reconstruction error, thus addressing the problem of 
subspace clustering in an end-to-end learning paradigm.
To achieve our goal, we propose a scheme to update subspaces within a deep neural network. 
This in turn frees us from the need of having an affinity matrix to perform clustering.
Unlike previous attempts, our method can easily scale up to large datasets, making it 
unique in the context of unsupervised learning with deep architectures.
Our experiments show that our method significantly improves the clustering accuracy while enjoying cheaper memory footprints. 

\end{abstract}

\section{Introduction}

Subspace Clustering (SC) is the de facto method in various clustering tasks such as
 motion segmentation~\cite{kanatani2001motion,elhamifar2009sparse,ji2014robust,ji2016robust}, face clustering~\cite{ho2003clustering,elhamifar2013sparse} and image segmentation~\cite{yang2008unsupervised,ma2007segmentation}. As the name implies, the underlying assumption in SC is 
that samples forming a cluster can be adequately described by a subspace. Such data modeling is natural in many applications.
One prime example is face clustering in which it has been shown that the face images of one subject obtained with a fixed pose and varying illumination lie in a low-dimensional subspace~\cite{lee2005acquiring}.

Most recent subspace clustering methods~\cite{elhamifar2013sparse,liu2013robust} assume that data points lie on a union of linear 
subspaces and construct an affinity matrix for spectral clustering. Although promising results on certain datasets are obtained, 
the performances degrade significantly when non-linearity arises in the data. Moreover, constructing the affinity matrix 
and performing clustering demand hefty memory footprints and processing power. 
To benefit from the concept of SC and its unique features,  two issues should be addressed;

\textbf{Non-Linearity:} Majority of the SC algorithms target clustering with linear subspaces. This is a very bold assumption 
and can hardly be met in practice. Some studies~\cite{patel2014kernel,yin2016kernel,xiao2016robust,ji2017low}
 employ kernel methods to alleviate this limitation. Nevertheless, 
kernel methods still suffer from the scalability issues~\cite{you2016scalable}. To make things more complicated, there is no guideline as how to choose the kernel function and its parameters which truly well-suited to subspace clustering. 

\textbf{Scalability:} With the current trend in analyzing big data, SC algorithms should be able to deal with large volume of data. 
However, most of the state-of-the-art methods for SC make use of an affinity matrix along norm regularization (\eg, $\ell_1$~\cite{elhamifar2009sparse,elhamifar2013sparse}, $\ell_2$~\cite{ji2014efficient} or nuclear ~\cite{liu2013robust,vidal2014low}). Not only building an affinity matrix demands for solving large scale optimization problems, but also performing spectral clustering on an affinity matrix, whose size is dictated by the number of samples, is overwhelming. 

In this paper, instead of constructing the affinity matrix for spectral clustering, we revisit the $k$-subspace clustering ($k$-SC) method~\cite{bradley2000k,tseng2000nearest,agarwal2004k} to design a novel and scalable method. In order to handle non-linear subspaces, we propose to utilize deep neural networks to project data to a latent space where $k$-SC can be easily applied. Our contributions in this paper are three-folds:

\begin{enumerate}
\setlength\itemsep{0.1em}
\item[-] We bypass the steps of constructing an affinity matrix and performing spectral clustering, which have been used in mainstream subspace clustering algorithms, and accelerate the computation by using a variant of $k$-subspace clustering. 
As a result, our method can handle datasets that are orders of magnitudes larger than those considered in traditional methods.
\item[-] In order to address non-linearity, we equip deep neural networks with subspace priors.  This in return enables us to learn an explicit non-linear mapping of the data that is well-suited for subspace clustering.
\item[-] 
We propose novel strategies to update subspace bases. When the size of the dataset at hand is manageable, we update subspaces in closed-form using Singular Value Decomposition (SVD) with a simple mechanism to rule out outliers. For large datasets, we update subspaces by making use of the stochastic optimization methods on the Grassmann manifolds.
\end{enumerate}

Empirically, evaluations on relatively large datasets such as  MNIST and Fashion-MNIST dataset~\cite{xiao2017/online} show that our proposed method achieves the state-of-the-art results in terms of  clustering accuracies and speed. 
\section{Related Work}
Linear subspace clustering methods can be classified as algebraic algorithms,  iterative methods, statistical methods and spectral clustering-based methods~\cite{vidal2011subspace}. Among them, spectral clustering-based methods~\cite{elhamifar2009sparse,liu2013robust,ji2014efficient,ji2014null,ji2015shape,you2016oracle} have become dominant in the literature. 
In general, spectral clustering-based methods solve the problem in two steps: encode a notion of similarity between pairs of data points into an affinity matrix; then, apply normalized cuts~\cite{shi2000normalized} or spectral clustering~\cite{ng2001spectral} on this affinity matrix. 
To construct the affinity matrix, recent methods tend to rely on the concept of {\it self-expressiveness}, which seeks to express each point in a cluster as a linear combination of other points sharing some common notions (\eg, coming from the same subspace).

The  literature on true end-to-end learning of subspace clustering is surprisingly limited. Furthermore and to the best of our knowledge, none of the deep algorithms can handle medium size datasets, let aside the large ones\footnote{Among all the datasets that have been tested, COIL100 with 7,200 images seems to be the largest one.}. 
In hybrid methods such as~\cite{peng2016deep}, hand-crafted features (\eg, SIFT~\cite{lowe2004distinctive} or HOG~\cite{dalal2005histograms}) are fed into a deep auto-encoder with a sparse subspace clustering (SSC) prior. The final clustering is then obtained by applying k-means or SSC on the learned auto-encoder features. 
Instead of using hand-crafted features, Deep subspace clustering Networks (DSC-NET)~\cite{ji2017deep} employ the deep convolutional Auto-encoder to nonlinearly map the images to a latent space, and make use of a self-expressive layer between the encoder and the decoder to learn the affinities between all the data points. Through learning affinity matrix within the neural network, state-of-the-art results on several traditional small datasets are reported in~\cite{ji2017deep}.
Nevertheless, relying on the whole dataset to create the affinity matrix, DSC-NET cannot scale for large dataset.

The SSC by Orthogonal Matching Pursuit (SSC-OMP)~\cite{you2016scalable} is probably the only subspace clustering which could be considered as ``scalable''. The main idea is to replace the large scale convex optimization procedure with the OMP algorithm in constructing the affinity matrix. Having said this,  SSC-OMP 
makes use of spectral clustering and hence still fails to really push subspace clustering for large scale datasets.

$k$-Subspace Clustering~\cite{tseng2000nearest,agarwal2004k}, an iterative methods, can be considered as a generalization of $k$-means algorithm. 
$k$-SC shows fast convergence behavior and can handle both linear and affine subspaces explicitly. However, $k$-SC methods are sensitive to outliers and initialization.
Attempts to make $k$-SC methods more robust include the work of Zhang \etal~\cite{zhang2012hybrid} and Balzano \etal~\cite{balzano2012k}. 
In the former, best $k$ subspaces from a large number of candidate subspaces 
are selected using a greedy combinatorial algorithm~\cite{zhang2012hybrid} to make the algorithm robust to data corruptions.
Balzano \etal~ 
propose a variant of $k$ subspaces method named $k$-GROUSE which can handle the missing data in subspace clustering. 
However, the resulting methods seem not to producing competitive results compared to methods relying on affinity matrices.


In this paper, we propose $k$-Subspace Clustering($k$-SC) networks which incorporate $k$-SC into a deep neural network embedding. This lets us not only bypass the affinity construction and spectral clustering procedure, but also handle data points lying in non-linear subspaces.

\section{$k$-Subspace Clustering($k$-SC) Networks}
Our $k$-subspace clustering networks leverage on the properties of deep convolutional auto-encoder and the $k$-subspaces clustering. In this section we will discuss the $k$-subspace property and the whole framework in detail.
\begin{figure*}[!t]
\centering
\includegraphics[width=0.90\linewidth]{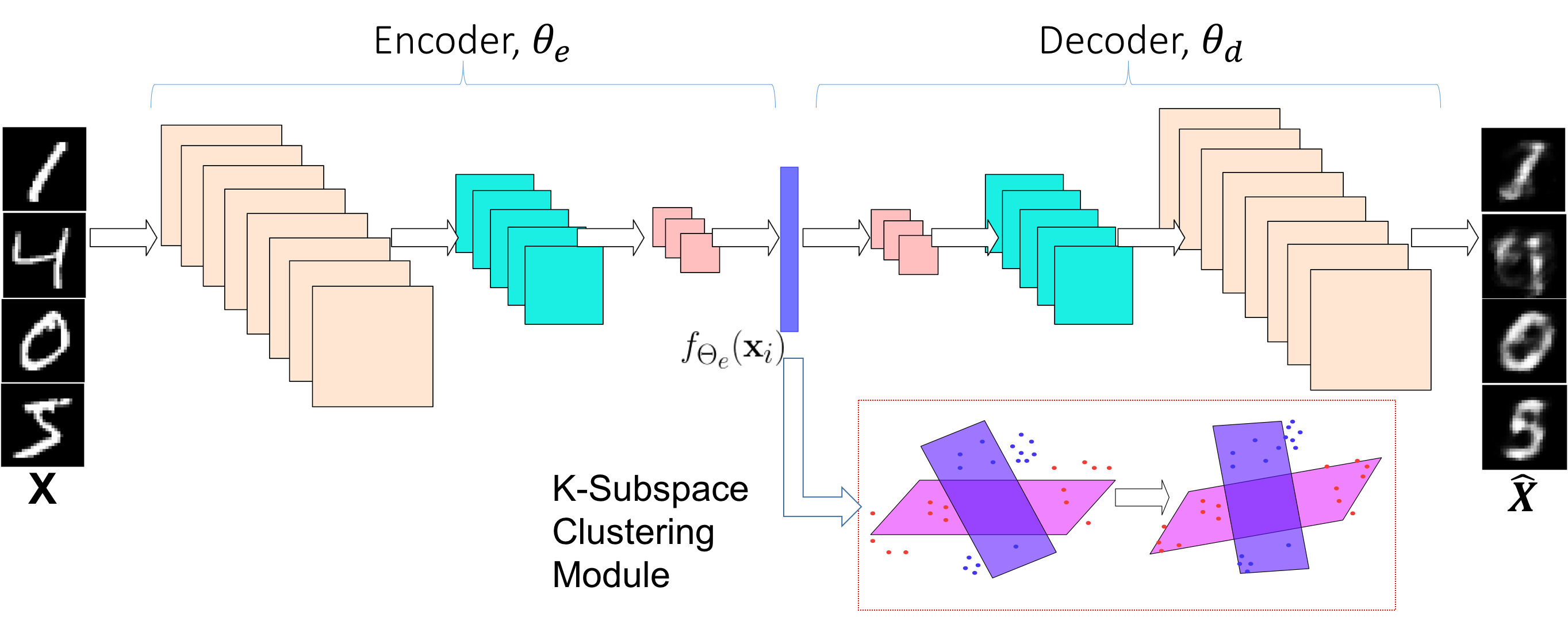}
\caption{Scalable deep $k$-subspace structure. As an example, we show our scalable batch-based subspace clustering with three convolutional encoder layers and three deconvolutional decoder layers. During the training, we first pre-train the deep convolutional auto-encoder by simply reconstructing the corresponding images, and then fine-tune the network using this pre-trained model as initialization. During the fine-tuning, the network  minimizes the sum of distances of each sample in the latent space to its closet subspace.} \label{fig:structure}
\end{figure*}
\subsection{$k$-Subspace Clustering}

Consider a collection of points $\{\mathbf{x}_1, \mathbf{x}_2, \cdots , \mathbf{x}_n\} \in \mathbb{R}^d$ belonging to a union of $k$ subspaces $\Mat{S}_1, \Mat{S}_2, \cdots , \Mat{S}_k$ of dimensions $p_1, p_2, \cdots , p_k$, respectively~\footnote{We assume $p=p_i,\forall i$ in the remainder.}. With slight abuse of notation, we will use $\Mat{S}_i$ to represent the basis of the subspace index by $i$, that is $\Mat{S}_i \in \mathbb{R}^{d \times p_i}$ and $\Mat{S}_i^\top \Mat{S}_i = \mathbf{I}_{p_i}$ with $\mathbf{I}_{p}$ denoting $p \times p$ identity matrix. The goal of subspace clustering is to learn the subspaces and assign points to their nearest subspaces. 
Once every data point is assigned to a subspace, the corresponding subspace basis can be re-calculated by SVD (will be shown shortly). Different from self-expressiveness-based methods which obtain the affinity matrix by solving large-scale optimization problems, $k$-SC seeks to minimize the sum of residuals of points to their nearest subspaces. The cost function of $k$-SC can be written as
\begin{equation}
\begin{split}\label{KSS}
& \min_{\{\mathbf{S}_i\}, \{w_{ij}\}} \sum_j^n\sum_i^k w_{ij}  \|\mathbf{x}_j - \mathbf{S}_{i}\mathbf{S}^\top_i \mathbf{x}_j\|_2^2, \\
& {\rm s.t.}  \quad  w_{ij} \in \{0,1 \} \quad  {\rm and} \quad \sum_{i=1}^k w_{ij} = 1\;.
\end{split}
\end{equation}

Given the subspace basis $\{\Mat{S}_1,\cdots,\Mat{S}_n\}$, the optimal value for $w_{ij}$ can be written as
\begin{equation}\label{member}
w_{ij} = 
\begin{cases}
1 \quad {\rm if} \quad  i = \argmin\limits_m \| \mathbf{x}_j - \Mat{S}_{m}\Mat{S}^\top_m \mathbf{x}_j\|_2^2\;,       \\
0 \quad {\rm otherwise}\;.
\end{cases}
\end{equation}


For the sake of discussion, let us arrange $w_{ij}$ into a membership matrix $\mathbf{W} \in \mathbb{R}^{k \times n}$. 
Beginning with an initialization of $k$ candidate subspaces bases, $k$-SC updates the membership assignments $w_{ij}$ and subspaces in an alternating fashion: 1) cluster points by assigning the nearest subspace as in Eqn.~\eqref{member}; 2) re-estimate the new subspace bases by performing SVD on the points of each cluster (the columns of $\mathbf{W}$ where the i-th row is 1). 
Similar to $k$-means, the whole algorithm works in an Expectation Maximization (EM) style, and is guaranteed to converge to a local minimum in a finite number of iterations. We will shortly show how stochastic optimization techniques can be applied to minimize the problem depicted in~\eqref{KSS}, equipping our solution with the ability to handle large-scale data.

\subsection{$k$-SC with Convolutional Auto-Encoder Network}
Denoising fully-connected Auto-Encoders (AEs) are widely used with generic clustering algorithms~\cite{pmlr-v70-yang17b,xie2016unsupervised}. 
We have found such structures difficult to train (due to the large number of parameters in the fully-connected layers) and propose to use convolutional AEs to learn the embeddings for $k$-SC.

Specifically, let $\theta$ denote the AE parameters, which can be decomposed into encoder parameters $\theta_e$ and decoder parameters $\theta_d$. Let $f_{\theta_e}(\cdot)$ be the encoder mapping function and $g_{\theta_d}(\cdot)$ as the decoder mapping function, both of which are composed of a sequence of convolution kernels and nonlinear activation functions. Our overall loss can be written as 
\begin{equation}
\begin{split}\label{cost}
\ell(\theta, \{\mathbf{S}_i\}, \mathbf{W}) = \ell_\mathrm{ae}(\theta) + \lambda \ell_\mathrm{ksc}(\{\mathbf{S}_i\},\mathbf{W})\;,
\end{split}
\end{equation}
where $\lambda$ is a regularization parameter to balance the reconstruction loss and the k-subspace clustering loss. The  auto-encoder reconstruction loss $\ell_{ae}$ is  defined as
\begin{equation}
\begin{split}\label{recon}
\ell_\mathrm{ae}(\theta) = \sum_j \|\mathbf{x}_j - g_{\theta_d}(f_{\theta_e}(\mathbf{x}_j))\|_2^2.
\end{split}
\end{equation}
The $\ell_\mathrm{ksc}(\theta)$ is the loss for subspace clustering and is written as
\begin{equation}
\begin{split}\label{ksc_ae}
 &\ell_\mathrm{ksc}(\{\mathbf{S}_i\},\theta) = \sum_{i,j} w_{ij}\| f_{\theta_e}(\mathbf{x}_j)-\mathbf{S_i}\mathbf{S_i}^\top f_{\theta_e}(\mathbf{x}_j)\|_2^2 \\
 &{\rm s.t.}\quad \Mat{S}_i \in \mathcal{G}(d,p), \; w_{ij} \in \{0,1 \} \;, \; \sum_{i=1}^k w_{ij} = 1,\; \forall ij\;,
\end{split}
\end{equation}
where $\mathcal{G}(d,p)$ denotes the Grassmann manifold consisting of $p$-dimensional subspaces with ambient dimension $d$.

As a pre-processing step, some of  traditional algorithms such as~\cite{you2016scalable,zhang2012hybrid} use PCA to project images onto a low-dimensional space. However, the mapping by PCA projection is linear and fixed. By contrast, our encoder function $f_{\theta_e}$ can update its parameters to adapt to a space which is subspace-clustering-friendly.

\section{Optimization}
\begin{algorithm}
\caption{Scalable $k$-Subspace Clustering (SVD update) }\label{alg:svd}
\begin{algorithmic}
\State \textbf{Input:} dimensionality of subspaces $p$, number of class K, epochs number T, batch size $b$ and dataset $\mathbf{x}_j$, $j=1,\cdots,N$
\State Pre-train CAE using $\mathbf{x}_j$, $j=1,\cdots,N$
\State Generate $ \{  \Mat{S}_i \} $ based on the pre-trained model and initial cluster labels
\State \textbf{for} $m = 1:T$ \textbf{do}
\State \quad \textbf{for} $ n = 1:b:N $
 \State \qquad Update the CAE parameters $\theta$ by Eqn.~\eqref{bp}
\State \quad \textbf{end}
 \State \quad Recalculate the latent space for the whole data set, 
 \State \quad $\mathbf{z}_j = f_{\theta_e}(\mathbf{x}_j), j = 1 ...N $
 \State \quad Assign the membership for every $\mathbf{z}_i$ as Eqn.\eqref{member} and 
 \State \quad rule out the farthest $10\%$ points as outliers, for each
 \State  \quad class we have set $\mathbf{Z}_i$
 \State \quad Update each subspace $ \Mat{S}_i$ through SVD decomposition on the $\mathbf{Z}_i$
 \State \textbf{end}
 \State \textbf{Output:} Subspaces \{$\Mat{S}_i$\}, and membership assignment $w_{ij}$ 

\end{algorithmic}
\end{algorithm}

The cost function~\eqref{cost} is highly non-convex and three sets of variables (\ie, $\mathbf{W},\theta$, and $\{ \mathbf{S}_i \}$) should be updated alternatively.
It is known that alternating optimization problems are not without difficulties. A strategy such as wake-and-sleep is a common practice to update one set of variables while fixing the others. 
As mentioned before, we first pre-train a CAE without having any information about  $\mathbf{W}$ and $\mathbf{S}_i$. 
Therefore, it is natural to obtain an initial state for $\mathbf{W}$ and $\{ \mathbf{S}_i \}$ directly from the output of the pre-trained CAE.
This is exactly how we initialize $\mathbf{W}$ and $\{ \mathbf{S}_i \}$. 

As shown in Fig~\eqref{fig:structure}, the gradient of the encoder comes from the loss of reconstruction and the loss of $k$-subspace clustering loss, \ie, 
\begin{equation}
\begin{split} \label{bp}
 \nabla_{\theta_e}\ell = \frac{\partial \ell_\mathrm{ae} }{\partial \theta_e}+ \lambda \frac{\partial \ell_\mathrm{ksc} }{\partial \theta_e}.
\end{split}
\end{equation}
By fixing $\{ \mathbf{S}_i \}$, the assignments $\mathbf{W}$ for a mini-batch can be obtained easily and the required gradient for updating 
the CAE follows by back-propagating the error. 
The most difficult part in our problem is to find a way to update the subspaces efficiently and accurately. Here we will explain two 
approaches to update the subspaces. The first method is based on the SVD decomposition and the second one makes use of the Riemannian geometry of Grassmannian to update the subspaces

\subsection{SVD Update}
\label{subsec:svd_update}

Although SVD decomposition is computationally more expensive, we empirically observe that the SVD can provide satisfactory results. 
In our optimization, we update the encoder through back-propagation, batch by batch, and update the subspaces by employing the SVD once per epoch. This is mainly because updating subspaces more frequently hinders the convergence. 
Intuitively, if the gradient takes the network to a bad direction, updating subspaces accordingly 
could intensify the negativity and worsen the CAE. Empirically, we observe updating subspaces after every epoch can neutralize the good and the bad directions of the gradient, yielding a stable framework. 

The outliers  may affect the subspace clustering badly, especially for $k$-subspace clustering. Therefore, when updating each subspace, we rule out the farthest $10\%$ points as outliers. That is, after back propagation on CAE, we pass all the data through the encoder and assign their membership. We then sort the distance between each sample and the subspace it belongs to, and remove the outliers. Finally, we apply SVD on the remainder of points assigned to a subspace to obtain its new basis. Note that we only need to compute the largest $p$ singular values and corresponding vectors to update a subspace. Specifically, after fixing $w_{ij}$ and $\theta$ in Eqn.~\eqref{ksc_ae}, updating the subspace basis $\Mat{S}_i$ translates to solving the following problem
\begin{equation}
\label{update_Si}
\argmin\limits_{\Mat{S}_i\in\mathcal{G}(d,p)}\|\Mat{Y}_i - \Mat{S}_i\Mat{S}_i^\top\Mat{Y}_i\|_F^2\;,
\end{equation}
where $\Mat{Y}_i$ consists of $\{f_{\theta_e}(\mathbf{x}_j)\}$ (as columns) that belong to cluster $i$. The solution to \eqref{update_Si} corresponds to the column space $\Mat{Y}_i$, which can be obtained by applying SVD on $\Mat{Y}_i$ and taking the top $p$ left singular vectors.





\subsection{Gradient based update}

If more frequent updates are required, the SVD solution can be replaced by a Riemannian gradient descent method based on the geometry of Grassmannian. In particular, let $\nabla \ell_{ksc}(\Mat{S}_i)$ be the gradient of the loss with respect to $\Mat{S}_i$ after an iteration (or 
accumulated gradient after a few iterations). In Riemannian optimization, $\Mat{S}_i$ is updated according to the following rule;
\begin{align}
\Mat{S}_i^{(t+1)} = \Upsilon_{\Mat{S}_i^{(t)}}\Big(-\eta \Pi_{\Mat{S}_i^{(t)}}\big(\nabla \ell_{ksc}(\Mat{S}_i) \big)\Big)\;.
\label{eqn:riem_update}
\end{align} 

\begin{figure}[!t]
\centering
\includegraphics[width=0.3\linewidth]{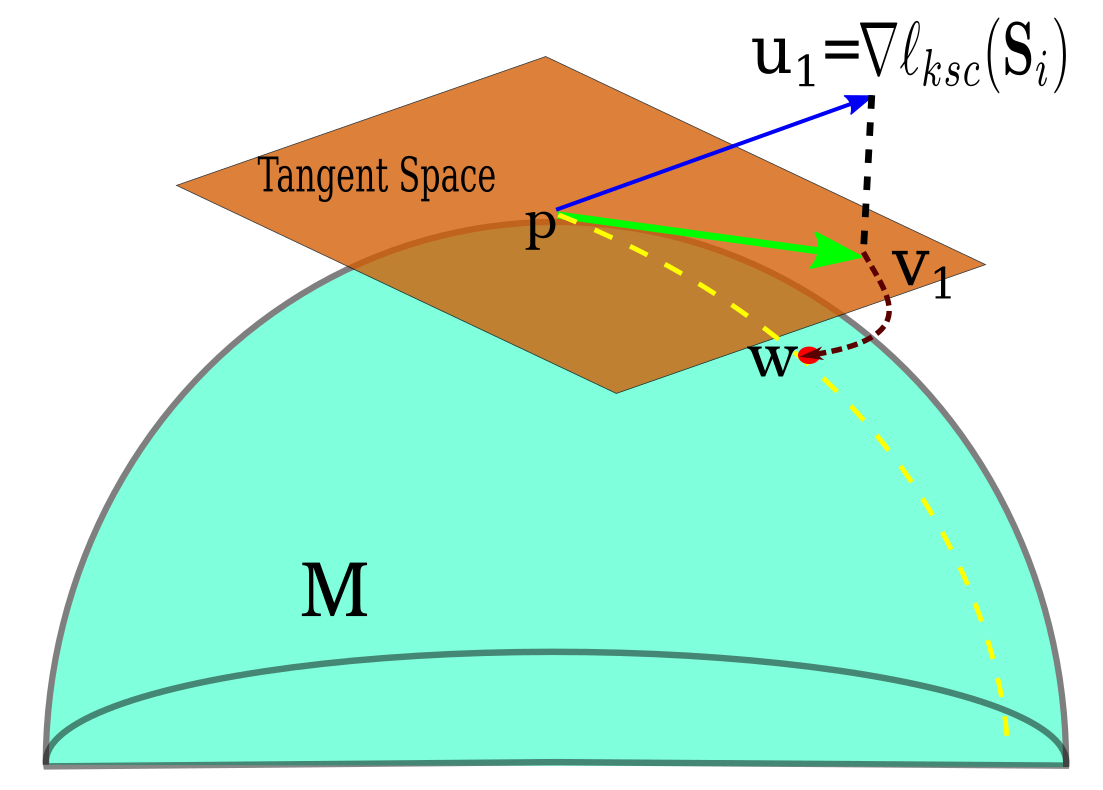}
\caption{ Illustration of how we update the gradient and keep the subspaces lie on the manifold} \label{fig:manifolds}
\end{figure}

We explain Eqn.~\eqref{eqn:riem_update} with the aid of Fig.~\ref{fig:manifolds}. First we note that a global coordinate system on 
a Riemannian manifold cannot be defined. As such Riemannian techniques make extensive use of the tangent bundle of the manifold to 
achieve their goal. Note that moving in the direction of $\nabla \ell_{ksc}(\Mat{S}_i)$ will take us off the manifold.
For a Riemannian manifold embedded in a Euclidean space (our case here), an ambient vector such as $\nabla \ell_{ksc}(\Mat{S}_i)$ can be 
projected orthogonally on the tangent space at the current solution $\Mat{S}_i^{(t)}$. We denote this operator by $\Pi$ in 
Eqn.~\eqref{eqn:riem_update}. The resulting tangent vector shown by the green arrow in Fig.~\ref{fig:manifolds} identifies a 
geodesic on the manifold. Moving along this geodesic (sufficiently) will guarantee to decrease the loss while preserving the 
orthogonality of the solution. In Riemannian optimization, this is achieved by a retraction which is local approximation to 
the exponential map on the manifold. We denote the retraction in Eqn.~\eqref{eqn:riem_update} by $\Upsilon$. The only remaining bit is $\eta$ 
which is the learning rate. For the Grassmannian, we have
\begin{align}
\label{eqn:proj_grassmann}
\Pi_{\Mat{S}} \big(\Vec{u}\big) &= \Big(\mathbf{I}_d - \Mat{S}\Mat{S}^\top\Big)\Vec{u}\;,\\
\label{eqn:retraction_grassmann}
\Upsilon_{\Mat{S}} \big(\Vec{u}\big) &= \mathrm{qf} \Big(\Mat{S} + \Vec{u}\Big)\;,
\end{align} 

In Eqn.~\eqref{eqn:retraction_grassmann}, $\mathrm{qf}$ is the Q part of the QR decomposition which is much faster than SVD.
Although the SVD can perform good enough in experiments, we provide the other method which is  faster in order to deal with very large datasets. 

\begin{algorithm}
\caption{Scalable $k$-Subspace Clustering(Gradients update) }\label{alg:grad}
\begin{algorithmic}
\State \textbf{Input:} dimensionality of subspaces $p$, number of class K, initial $ \{  \Mat{S}_i \} $, epochs number T, pre-trained CAE, batch size $b$ and dataset $\mathbf{x}_j$, $j=1,...,N$
\State\textbf{for} $m = 1:T$ \textbf{do}
\State \quad \textbf{for} $n = 1:b:N $
\State \qquad Assign the membership for every point based on the distance
\State \qquad to each subspace as Eqn.~\eqref{member}
\State \qquad Compute the gradients with respect to each subspace as 
\State \qquad Eqn.~\eqref{ksc_ae}  $\ell_{ksc}(\Mat{S}_i)$  and store them
\State \qquad Update the CAE parameters $\Theta$ by Eqn.~\eqref{bp}
\State \quad \textbf{end}
\State \quad Project the gradients on Grassmannian manifolds 
\State \quad based on  Eqn.~\eqref{eqn:proj_grassmann}
\State \quad Apply the gradients on the corresponding subspace 
\State \quad accord to Eqn.~\eqref{eqn:retraction_grassmann}
\State \textbf{end}
\State \textbf{Output:} subspaces  $ \{  \Mat{S}_i \} $, and membership assignment $w_{ij}$
\end{algorithmic}
\end{algorithm}

\section{Experiment}
We use Tensorflow~\cite{abadi2016tensorflow} to build our networks. 
We used MNIST dataset~\cite{lecun1998gradient} in our first experiment. MNIST is not considered as a standard dataset for previous subspace clustering algorithms, since the size of this dataset is far beyond the size that traditional algorithms can handle. In addition, the original images do not follow the structure of linear subspaces. Taking advantage of CAE with our $k$-subspace clustering module, we aim to project all the MNIST data into a space which is more friendly for subspace clustering. In order to enforce our conclusion, we also evaluate our method on Fashion-MNIST dataset~\cite{xiao2017/online}, a similar dataset to MNIST but with fashion images. Fashion-MNIST has 10 classes, with image being gray scale and of size  $28 \times 28$. The images of Fashion-MNIST come from the fashion products which are classified based on a certain assortment and manually labeled by in-house fashion experts and reviewed by a separate team. It contains more variations within each class and it is thus more challenging compared to MNIST.

\subsubsection{Baseline Methods}


For most of the baselines and our method, we evaluate them on the whole datasets of MNIST and Fashion-MNIST with all 70000 images (including both training and testing sets).
We compare our solution with the following generic clustering algorithms:

1) \textrm{$k$-Means}~\cite{lloyd1982least}: $k$-means finds clusters based on spatial closeness. As an EM method, it heavily relies on good initialization. Hence, for $k$-means (and other $k$-means based methods), we run the algorithm 20 times with different centroid seeds and report the best result.

2) \textrm{Deep Embedded Clustering (DEC)}~\cite{xie2016unsupervised}: A rich structure for the MNIST dataset is proposed in~\cite{xie2016unsupervised} which we follow here. In particular, stacked autoencoder(SAE)~\cite{bengio2007greedy} along layer-wise pre-training was considered. 
The structure of the network reads as 784-500-500-2000-10. 
Image brightness is scaled from 0-1 to 0-5.2 
to boost the performance. We observe that this method is highly sensitive to network parameters in the sense that even a small change in the structure will result in a significant performance drop. However, the feature extracted by the pre-trained model is very discriminative, \ie, even simply using k-means on top of it can achieve competitive results. We call the feature extracted by this network the SEA features in the sequel.

3) \textrm{Deep Clustering Network (DCN)~\cite{pmlr-v70-yang17b}}: Based on the vanilla SAE, Yang \etal  propose to add $k$-means clustering loss in addition to the data reconstruction loss of SAE. 

4) \textrm{Stacked Auto-Encoder followed by $k$-Means (SAE-KM)}: Extract features with SAE followed by applying $k$-means. 

5) \textrm{PCA followed by $k$-subspace (PCA-KS)}: It projects the original data onto a low-dimensional space at first, then use $k$-subspace to obtain the final results. Since PCA is a linear projection, it helps the readers to understand where the improvements come from compared to our nonlinear projection. The results are reported based on the 10 trails due to the randomness of initialization when employing $k$-subspace.

6) \textrm{Convolutional Auto-Encoder followed by $k$-Means (CAE-KM)}: Extract features with SAE and then apply $k$-means. This is also the initialization for our method. It also can be considered as an evaluation of the quality of our initialization.   


For those subspace clustering algorithms that rely on affinity matrix construction and spectral clustering, since they are not scalable to the whole dataset, we can report their results on the test sets (with 10000 images) only. We list several state-of-the-art subspace clustering algorithms for baselines: \textrm{Sparse Subspace Clustering (SSC)}~\cite{elhamifar2013sparse}, \textrm{Low Rank Representation (LRR)}~\cite{liu2013robust}, \textrm{Kernel Sparse Subspace Clustering (KSSC)}~\cite{patel2014kernel}, \textrm{SSC by Orthogonal Matching Pursuit(SSC-OMP)}~\cite{you2016scalable} and the latest one \textrm{Deep Subspace Clustering Networks} (DSC-Net)~\cite{ji2017deep}.

\subsubsection{Evaluation Metric}
For all quantitative evaluations, we make use of the unsupervised clustering accuracy rate, defined as
\begin{equation}
{\rm ACC }\;\% =\max_M \frac{\sum_{i=1}^n \mathbf{1}(l_i = M(s_i))}{n}\times 100\%\;.
\end{equation}
where $l_i$ is the ground-truth label, $s_i$ is the subspace assignment produced by the algorithm, and $M$ ranges over all possible one-to-one mappings between subspaces and labels. The mappings can be efficiently computed by the Hungarian algorithm.
 We also use normalized mutual information (NMI) as the additional quantitative standard. NMI scales from 0 to 1, where a smaller value means less correlation between predict label and ground truth label. Another quantitative metric is the adjusted Rand index (ARI), which is scaled between -1 and 1. The larger the ARI, the better the clustering performance. 
  
\subsubsection{Implementation}
We build our CAE in a bottle-neck structure, meaning we decrease the number of channels and the size of feature maps layer by layer. We design a six layer convolutional auto-encoder, where the kernel size in the first layer is $5 \time 5$ and in the last two layers of the encoder is $3 \times 3$. We set the number of channels in each layer to $20-10-5$ for the encoder, and the reverse for the decoder since they are symmetric in structure. Between layers, we set the stride to 2 in both horizontal and vertical directions, and use rectified linear unit (ReLU) as the non-linear activations. We use the same structure for both MNIST and Fashion-MNIST datasets. 

Instead of greedy layer-wise pre-training~\cite{pmlr-v70-yang17b,xie2016unsupervised}, we pre-trained our network end-to-end from random initialization, until the reconstructed images are similar to the input ones (200 epochs suffice for pre-training). For subspaces initialization, we randomly sampled 2000 images and use DSC network to generate the clusters and corresponding subspaces. We noticed that initialization by the DSC subspaces leads to a model that under-performs 
in the beginning as compared to the $k$-Means algorithm. Nevertheless, our algorithm successfully recovers from such an initialization in all the experiments.
During the optimization we use Adam~\cite{kingma2014adam} optimizer, an adaptive momentum based gradient descent method, to minimize the loss, where we set the learning rate to $1.0\times 10^{-3}$ in both our pre-training and fine-tuning stages. For different datasets, the only two parameters need tuning are the $\lambda$ in~\eqref{cost} and the subspace ambient dimension $n$ , since the subspace intrinsic dimension $p$ is fixed by the number of feature map of CAE.

\subsection{MNIST Dataset}

In this section, we will report and discuss results on the MNIST dataset. 
To the best of our knowledge, existing subspace clustering methods, with raw images as input, have not achieved satisfactory results on this dataset. As far as we know, the best performance reported in~\cite{peng2017deep} is in the range $58\%-65\%$, where the DSIFT features are employed.

On MNIST, we fix our subspace dimension as 7, which means each subspace lies on a Grassmannian manifold $\mathcal{G}(80,7)$.
The $\lambda$ is set to 0.08, which balances between subspace clustering and CAE data reconstruction. 
Table~\eqref{tab:MNIST:all} reports the results of all the baselines, including both subspace clustering algorithms and generic clustering algorithms. $k$SCN-S is to update the subspaces by employing the SVD decomposition, and $k$SCN-G
 stands for updating the subspaces by the Grassmannian gradients, which empirically is not  as stable as the SVD updating scheme, probably due to the stochastic nature of each gradient step.
This Grassmannian update, however, runs faster and takes less time to converge. We run our methods 15 times and report the average. The results of DEC are taken from the original paper. 
We tune the parameters for DCN very carefully and report the best results.

\begin{figure}[t]
\begin{center}
 \begin{tabular}{cc}
     \subfigure[CAE feature]{\includegraphics[width=0.28\linewidth]{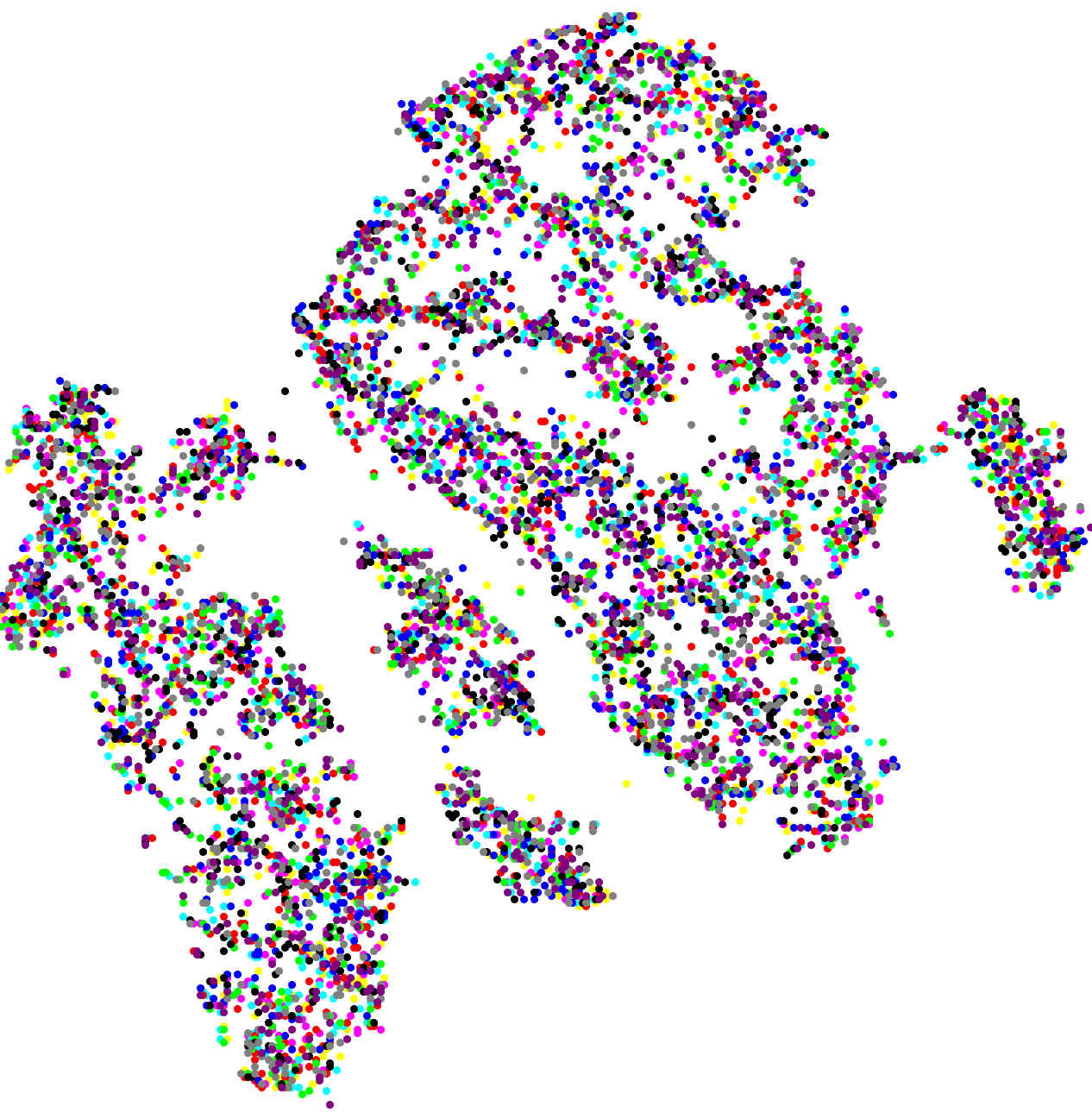}}  &\qquad\qquad\qquad
    \subfigure[Our latent space]{\includegraphics[width=0.28\linewidth]{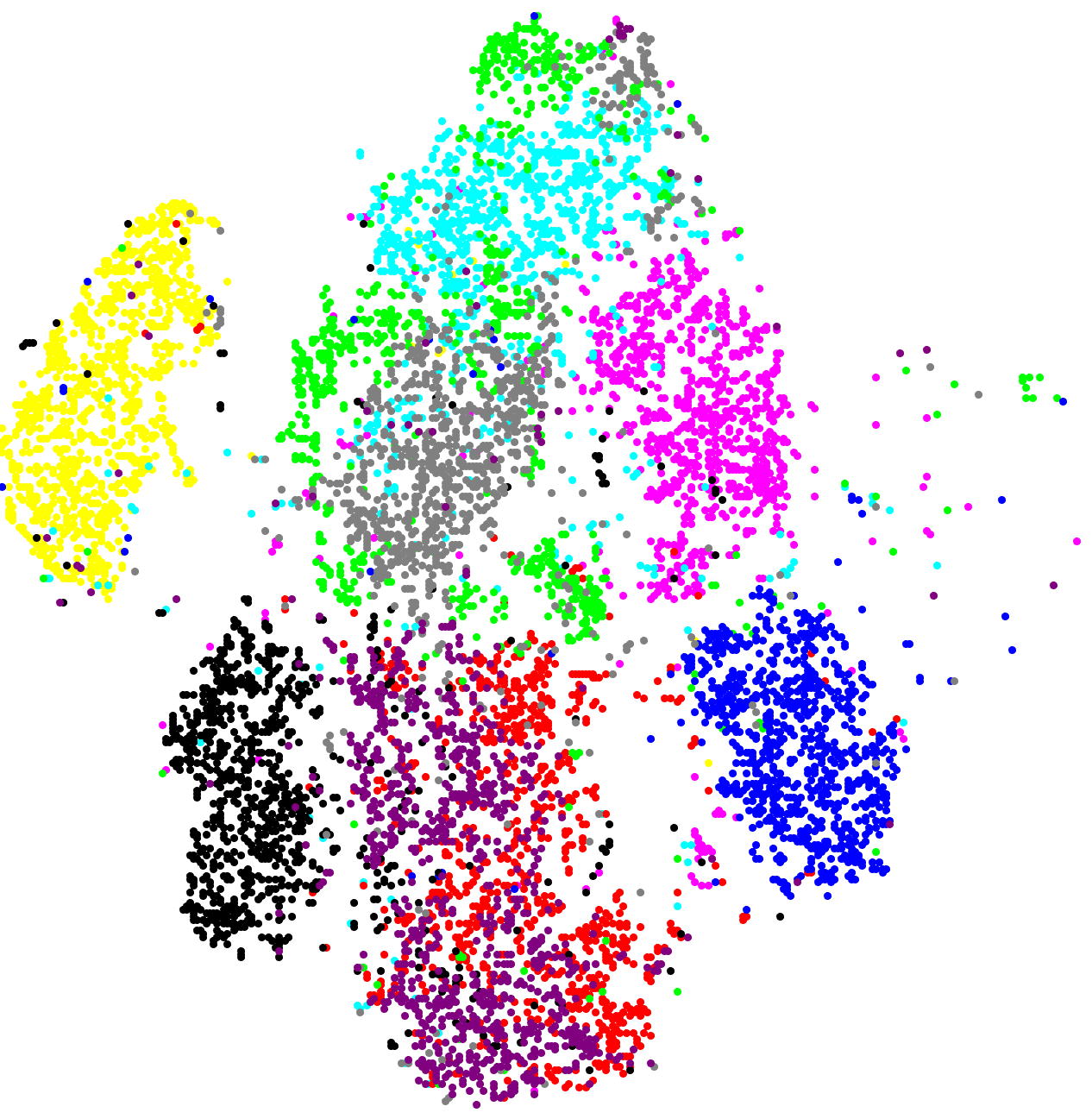}} 
	\\
    \end{tabular}
\caption{Visualization using t-SNE on the latent space generated by projecting the testing set images on pre-trained CAE and our network. Points marked with the same color belong to the same class.} \label{TSNE:MNIST}
\end{center}
\end{figure}

Among all the algorithms, our algorithm achieves the best performance in ACC and ARI. Especially for ACC, ours is $3 \%$ higher than the second best, namely DEC. From the results, it is not difficult to conclude that the DEC and DCN perform only marginally better than SAE-KM, which is the initialization for DEC and DCN. Specifically, DEC improvements over the initialization are around $3\%$ and DCN only boosts around $1.5 \%$ over SAE-KM. By contrast, our method starts from CAE-KM (with $51\%$ ACC), and improves it by $36.14\%$ to $87.14\%$ ACC. The improvement can be visualized by Fig~\eqref{TSNE:MNIST}, which shows the projections of CAE feature space and the latent space of our network in a two-dimensional space. Compared to CAE features, which are all mixed up, our latent space are well separated even though the two-dimensional space is not suitable for visualizing subspace structure as they reside in high-dimensional ambient space.

For traditional subspace clustering algorithms, around 37 Gigabytes of memory is required to store the affinity matrix, which is computationally prohibitive. Therefore, we contrast our algorithm against 
SSC, LRR, KSSC, SSC-OMP and Deep Subspace Clustering Networks on a smaller experiment, namely only using the 10000 test images of the MNIST dataset (see Table.~\eqref{tab:subs} for results). Note that SSC-OMP completely fails in dealing with feature generated by SAE and CAE, achieving around $12\%$ ACC and $2\%$ NMI. 
Generally speaking, with more samples, better accuracies are expected. 
We can see that all the subspace clustering algorithms using the SAE feature perform better compared to using 
CAE feature. To some extent, it proves that there exists a nonlinear mapping which is more favorable to subspace clustering. At the same time, our algorithm still achieves the best results within all subspace clustering algorithms, even higher that DSC-Net. 


\begin{table}
\begin{center}
\caption{ Results on MNIST (70000 samples). \label{tab:MNIST:all}}
\begin{tabular}{c|c|c|c|c|c|c|c|c|c|}
\hline
\multirow{1}{*}{} & \multicolumn{1}{c|}{SAE-KM} & \multicolumn{1}{c|}{CAE-KM} & \multicolumn{1}{c|}{K-means} & \multicolumn{1}{c|}{PCA-KS}& \multicolumn{1}{c|}{DEC}& \multicolumn{1}{c|}{DCN}& \multicolumn{1}{c|}{$k$SCN-G} & \multicolumn{1}{c|}{$k$SCN-S} \\
\hline
ACC&  81.29\% & 51 \%  & 53\% & 68.53\% & 84.3\%& 83.31\%& 82.22\% & \textbf{87.14\%}\\\hline
NMI & 73.78 \% & 44.87\% & 50 \% & 64.17\% & 80\% & \textbf{80.86}\% & 73.93\% & 78.15\%  \\ \hline
ARI & 67\% &  33.52 \%  & 37 \%  & 54.17\% & 75\% & 74.87\% & 71.10\% & \textbf{75.81} \%\\
\hline
\end{tabular}
\end{center}
\end{table}

\begin{table}
\begin{center}
\caption{The results of subspace clustering algorithms on the test sets of the MNIST and Fashion-MNIST datasets, the best two are in bold} \label{tab:subs}
\begin{tabular}{c|c|c|c|c|}
\multirow{2}{*}{} & \multicolumn{2}{c|}{MNIST} & \multicolumn{2}{c|}{Fashion-MNIST}\\
\cline{2-5}
\hline
 & ACC  & NMI &    ACC  & NMI \\
\hline
SSC-SAE &  75.49\%  &  66.26\% &  52.33 \%  &  51.26\% \\
SSC-CAE&  43.03 \%& 56.81\% & 35.31\%& 18.10\% \\
LRR-SAE& 74.09\% & 66.97\% &  \textbf{58.09}\% & 59.19\%   \\ 
LRR-CAE& 51.37\% & 66.59\% & 34.43\% &  18.57\%  \\
KSSC-SAE& \textbf{81.53}\% & \textbf{84.53}\% & 57.10\% & \textbf{60.40}\%  \\
KSSC-CAE& 56.42\%  &  65.66\%  & 35.41\%    & 18.18\% \\
\hline
DSC-Net &      53.20\% &  47.90\%        &   55.81\%     &   54.80 \%   \\
\hline
$k$SCN-S &  \textbf{83.30}\% & \textbf{77.38}\% & \textbf{60.02}\% & \textbf{62.30}\%    \\
\hline
\end{tabular}
\end{center}
\end{table}

\subsection{Fashion-MNIST}

Unlike MNIST dataset, which only contains simple digits, every class in Fashion-MNIST has different styles and come from different gender groups: men, women, kids and neutral. In Fashion-MNIST, there are 60000 training images and 10000 test images. In our case, 
we pre-trained and fine-tuned the network using the whole dataset. On Fashion-MNIST, we fix our subspace dimension to 11
and set $\lambda$ to 0.11. 

\begin{figure}[t]
\begin{center}
 \begin{tabular}{cc}
     \subfigure[CAE feature]{\includegraphics[width=0.28\linewidth]{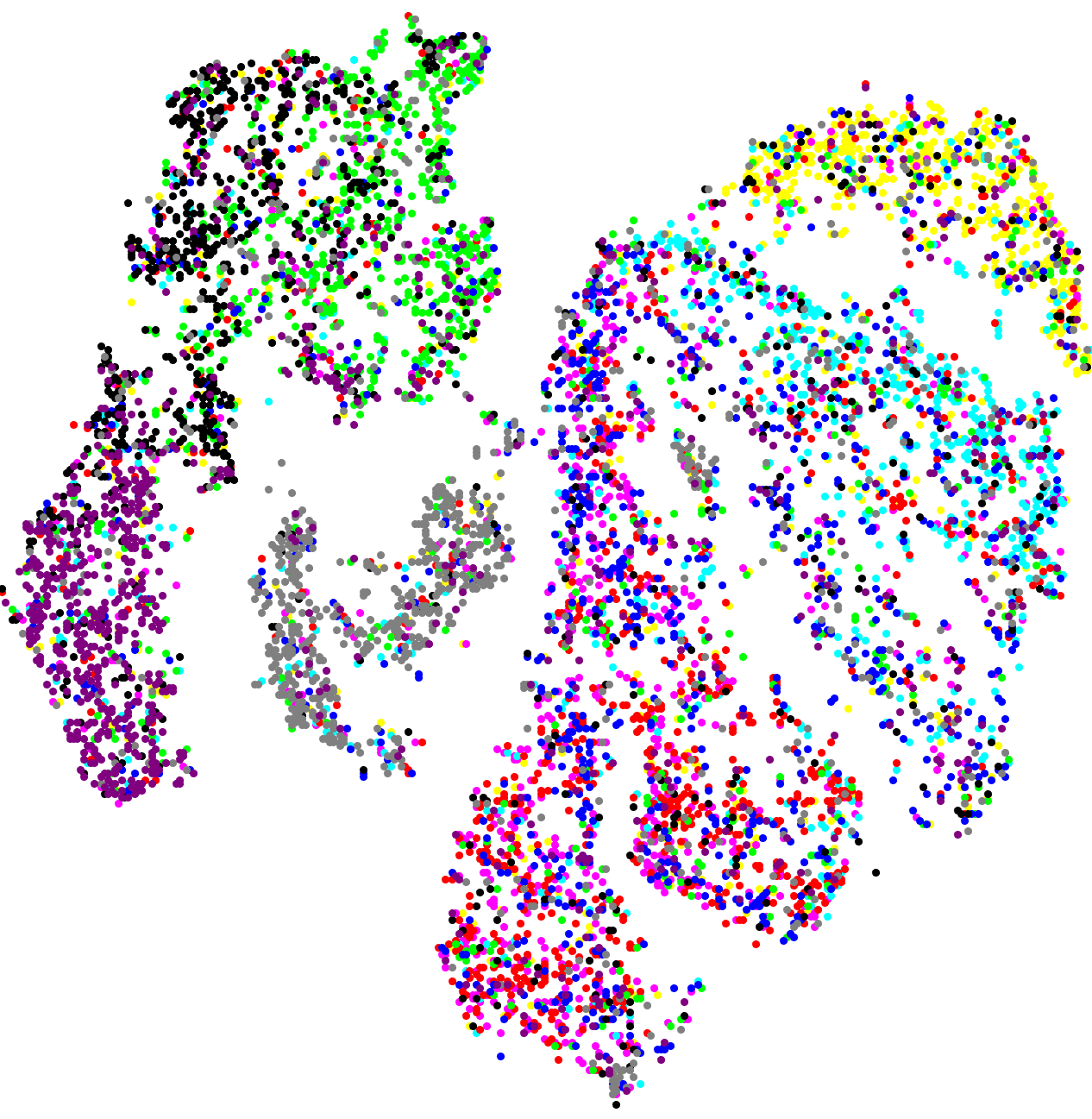}} &\qquad\qquad\qquad
    \subfigure[Our latent feature]{\includegraphics[width=0.28\linewidth]{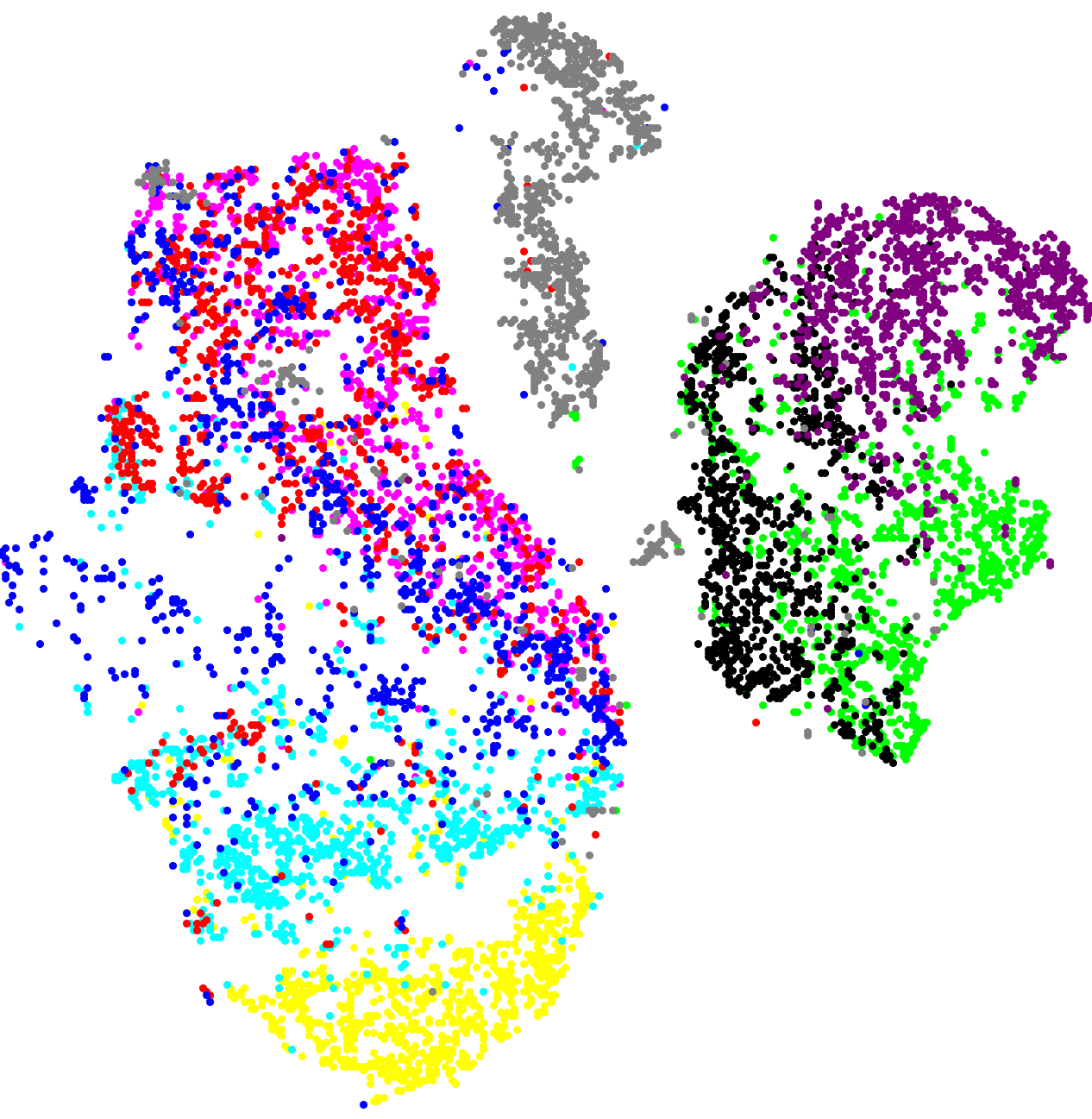}} 
	\\
    \end{tabular}
\caption{Visualization using t-SNE for the latent space generated by  pretrained CAE and our network on Fashion-MNIST. Points marked with the same color belong to the same class.} \label{TSNE:fashion}
\end{center}
\end{figure}

Consistent with the MNIST dataset, the DCN sightly improves upon its initialization (SAE-KM) in terms of ACC and NMI. Moreover, we find out that the DCN algorithm works better with smaller learning rates, which in turn requires more epochs to converge properly. From Table~\eqref{tab:fashion:all}, we can see that our method still improves the accuracy by $24\%$ compared to our initialization, and outperforms other algorithms. The t-SNE maps in Fig.~\eqref{TSNE:fashion} show that there exists a subspace structure in our latent space even in two dimensional space. 

Table~\eqref{tab:subs} shows that the subspace clustering algorithms also achieve acceptable results on the 10000 test sets, with our algorithm being the best among all. Compared to other subspace clustering algorithms, our algorithm runs much faster, only requiring less than 8 minutes
 (including pre-training and fine tuning with subspace clustering) to generate final results, whereas the traditional algorithms need at least 40 minutes to process these 10000 samples even after the dimensionality reduction.


\begin{table}
\begin{center}
\caption{Results on Fashion-Mnist \label{tab:fashion:all}}
\begin{tabular}{c|c|c|c|c|c|c|c|c|}
\hline
\multirow{1}{*}{} & \multicolumn{1}{c|}{SAE-KM} & \multicolumn{1}{c|}{CAE-KM} & \multicolumn{1}{c|}{K-means} & \multicolumn{1}{c|}{PCA-KS}& \multicolumn{1}{c|}{DCN}& \multicolumn{1}{c|}{$k$SCN-G} & \multicolumn{1}{c|}{$k$SCN-S} \\
\hline
ACC&  54.35\% & 39.84 \%  & 47.58\% & 53.41\% & 56.14\%& 58.67\%& \textbf{63.78}\% \\ \hline
NMI & 58.54 \% & 39.80\% & 51.24 \% & 57.5\% & 59.4\% & 52.88\% & \textbf{62.04}\%   \\ \hline
ARI &  41.86\% &  25.93 \%  & 34.86 \%  & 41.17\% & 43.04\% & 42\% & \textbf{48.04}\%   \\
\hline
\end{tabular}
\end{center}
\end{table}

\subsection{Further Discussion}

Based on the above experiments, we observe that our algorithm consistently achieves higher accuracies  as compared to DCN (even with the initialization using CAE). One may argue that the performance gain over DCN is due to the fact that unlike SAE, CAE can be trained easily\footnote{In our experiments, the number of parameters in SAE is 2600 times more than that of CAE.}. To verify that this is not the case, we replace the SAE with the CAE in DCN to see whether DCN can generate competitive results. Table~\eqref{tab:discussion} demonstrates that even with the CAE, the DCN cannot boost the clustering results as much as ours. On MNIST, DCN-CAE can hardly improve the accuracy and NMI; on Fashion-MNIST, it can increase the accuracy more than 3 percent (and NMI around 1 percent). 
This can be attributed to the loss introduced by $k$-means in DCN, compared to our $k$-subspace clustering loss which we believe is more robust. 
In other words, the subspace structure could be more desirable than cluster centroids in high dimensional space. 

\begin{table}
\begin{center}
\caption{The performance of the DCN-CAE and its CAE initialization.} \label{tab:discussion}
\begin{tabular}{c|c|c|c|c|}
\hline
\multirow{2}{*}{} & \multicolumn{2}{c|}{MNIST} & \multicolumn{2}{c|}{Fashion-MNIST}\\
\cline{2-5}
 & ACC  & NMI & ACC  & NMI \\
\hline
DCN-CAE  &  51.10 \%  &  45.18 \%& 45.64 \%& 47.8\% \\
Initilization& 50.98\% & 44.87 \% & 42.38  \% & 46.75  \%  \\ 
\hline
\end{tabular}
\end{center}
\end{table}

\section{Conclusions}
In this paper, we  proposed a scalable deep $k$-subspace clustering algorithm, which combined the $k$-subspace clustering and convolutional auto-encoder in a principle way. Our algorithm makes it possible to scale subspace clustering algorithms to large datasets. Furthermore, we proposed two efficient and robust schemes to update the subspaces. These allow our $k$-SC networks to iteratively fit every sample into its corresponding subspace and update the subspaces accordingly, even from a bad initialization (as observed in our experiments).

Our extensive experiments on MNIST and Fashion-MNIST dataset demonstrated that our deep $k$-subspace clustering method provides significant improvements over various state-of-the-art subspace clustering solutions in terms of clustering accuracy and efficiency. 


\bibliographystyle{splncs04}
\bibliography{deepksubspace.bib}

\end{document}